\documentclass{article}

\usepackage[sglblindworkshop, final]{neurips_2026}

\workshoptitle{Agents in the Wild (AIWILD)}
\usepackage[utf8]{inputenc} % allow utf-8 input
\usepackage[T1]{fontenc}    % use 8-bit T1 fonts
\usepackage{hyperref}       % hyperlinks
\usepackage{url}            % simple URL typesetting
\usepackage{booktabs}       % professional-quality tables
\usepackage{amsfonts}       % blackboard math symbols
\usepackage{nicefrac}       % compact symbols for 1/2, etc.
\usepackage{microtype}      % microtypography
\usepackage{xcolor}         % colors
\usepackage{cleveref}
\usepackage{graphicx}
\usepackage{subcaption}
\usepackage{color,soul}

\title{The Calls are Coming from Inside the Model: Investigating Probe-based Detection of Tool-Calling Errors in LLMs}

\author{%
  Eric Yeats\thanks{Corresponding Author}, Brendan Kennedy, Loc Truong, John Buckheit, Jung Lee, Jesse Friedbaum \\
  Pacific Northwest National Laboratory\\
  \texttt{\{first\}.\{last\}@pnnl.gov} \\
  \And
  John Emanuello \\
  National Security Agency \\
  \And
  Henry Kvinge \\
  Pacific Northwest National Laboratory \\
}

\begin{document}

\maketitle

\begin{abstract}
  The hidden states of large language models (LLMs) are known to capture rich information relating to model knowledge and behavior that can be hard to extract from examination of input and output alone. As LLM-based systems increasingly interface with the external world, one area of concern is detecting incorrect or improper use of tools. Motivated by this, we study the effectiveness of using linear probes to detect incorrect tool-calls, measuring probe efficacy across 18 tool-calling LLMs evaluated on the Berkeley Function Calling Leaderboard and 35k tool-calling instances from Glaive. Overall, we find that probing is an effective means to catch a range of different tool-calling errors, including errors arising from using an argument that has the wrong value but the correct type, which might not be recorded by standard logging frameworks. Important factors in success include model size, probing layer, and model post-training type. We also show that probes are capable of generalizing to novel types of errors, which is critical in real world deployments. 
\end{abstract}

\section{Introduction}

Large language models (LLMs) have demonstrated remarkable capabilities across a wide range of tasks, including the ability to interface with external tools and APIs through structured function calls \citep{achiam2023gpt, touvron2023llama}. This \textit{tool-calling} capability is increasingly central to agentic AI systems, where LLMs must select appropriate tools, supply correct arguments, and chain multiple calls to accomplish complex goals \citep{schick2023toolformer, qin2024toolllm}. As these systems are deployed in production settings, the reliability of tool-calls becomes critical.

LLM hallucinations, or outputs that are fluent but factually incorrect or unfaithful to context, remain pervasive and have received substantial attention in the broader NLP community \citep{huang2025survey, bang2023multitask}. A prominent line of work has shown that the internal representations, or \textit{hidden states}, of LLMs contain rich signals that are predictive of model errors and hallucinations \citep{burns2022discovering, kadavath2022language, yeats2026geometric}. These findings, along with the linear representation hypothesis \cite{park2023linear}, have motivated the development of lightweight \textit{probes} \cite{azaria2023internal,orgad2025llms}, linear classifiers or small MLPs trained on hidden states. Probes act as inexpensive monitors that can flag potential errors at inference time without requiring multiple evaluations of the LLM or access to external verifiers \citep{varshney2023stitch, chen2024inside}.

Despite this progress, hallucination detection via probing has been studied primarily in the context of factual question answering and text summarization. Tool-calling presents a distinct challenge: errors are structural and semantic rather than purely factual, and they span a diverse taxonomy \citep{patil2024gorilla, patil2025berkeley}. The field of tool-calling error detection with hidden states remains underexplored. The most closely related work to ours is \citet{healy2026internal}, who investigate probes trained on the \textit{last layer} of LLM hidden states for tool selection hallucination detection. They compare probes with consistency baselines, showing that probes are a lightweight and effective method for tool-call error prediction. However, their study is limited to just three models, does not investigate how probe effectiveness varies across different model sizes, architectures, and training regimes, and does not validate whether probing is detecting narrow failure patterns or is actually capturing more general notions of tool-calling incorrectness.

Our exploratory work extends this line of research in two key directions. First, we conduct a \textbf{large-scale study of probe effectiveness across 18 tool-calling LLMs} evaluated on the Berkeley Function Calling Leaderboard (BFCL) \citep{patil2025berkeley}, spanning model sizes from 1B to 70B parameters and a variety of post-training regimes. We identify consistent trends: model size is the strongest predictor of probe effectiveness, followed by recording layer depth and whether the model has been specifically fine-tuned for tool use. Second, we investigate the \textbf{generalization of probes to novel error types} by training probes on one subset of the error taxonomy and evaluating on a held-out subset. We find that probes trained on one tool-calling error type largely transfer to detecting others, and that this transfer improves with model scale, aligning with evidence that larger models develop more general and abstract internal representations of what it means to perform an `incorrect' tool-call. We include additional analysis using 35k tool-calling instances from the \href{https://huggingface.co/datasets/glaiveai/glaive-function-calling-v2}{\texttt{glaive-function-calling-v2}} dataset in the appendix (\Cref{app:glaive_experiments}).

\section{Methodology}
\label{sec:methodology}

In this study we extract the hidden state at the \textit{last token} of the generated tool-call. This token aggregates contextual information from the full input and has been shown to be effective for probing \citep{azaria2023internal, healy2026internal}. For a specified layer $\ell$ and a sequence of $T$ generated tokens corresponding to a tool-call, we extract the hidden state: $\mathbf{z}^{(\ell)} = \mathbf{h}^{(\ell)}_{T} \in \mathbb{R}^{d}$, yielding one vector per layer and per tool-call.

\paragraph{Linear Probes} A \textit{probe} is a lightweight classifier trained on frozen hidden states to predict tool-call correctness. Let $\mathcal{D} = \{(\mathbf{z}^{(\ell)}_i, y_i)\}_{i=1}^{N}$ denote a labeled dataset of $N$ tool-call hidden states at layer $\ell$, where $y_i \in \{0, 1\}$ indicates whether the $i$-th tool-call is correct ($y_i = 0$) or incorrect ($y_i = 1$).
We train a linear probe, an $\ell_2$-regularized logistic regression model, on $\mathcal{D}$, $p_\theta(y=1 \mid \mathbf{z}^{(\ell)}) =\sigma\!\left(\mathbf{w}^\top \mathbf{z}^{(\ell)} + b\right)$. Here $\sigma(\cdot)$ is the sigmoid function, and $\mathbf{w} \in \mathbb{R}^{d}$ and $b \in \mathbb{R}$ are learned parameters. The probe is fit by minimizing the $\ell_2$-regularized binary cross-entropy loss on the training split. Input features are standardized to zero mean and unit variance prior to training. Probes are trained independently for each recorded layer $\ell \in \{1, \ldots, L\}$, yielding a family of $L$ probes, whose detection efficacy across layer depth we analyze in \Cref{sec:results}.

\paragraph{Perplexity Baseline} As a baseline, we compute the perplexity of the generated tool-call. Given the prompt $x$ and a tool-call $y_{1:T}$, we compute $\mathrm{PPL} = \exp\!\big(-\frac{1}{T}\sum_{t=1}^{T}\log p(y_t \mid x, y_{<t})\big)$, where the average is taken over the tool-call tokens only and prompt tokens are excluded. Higher perplexity indicates lower model confidence, so we use it directly as the error score ($y=1$ is incorrect).

\paragraph{Metrics} We evaluate probe and perplexity performance using two threshold-independent metrics that do not require selecting a decision threshold and capture complementary aspects of detection quality: area under the receiver operating characteristic (AUROC) and area under the precision-recall curve (AUPR). AUPR results are available in the appendix, and we provide reviews of the definitions of these in \Cref{sec:metrics}. We also provide FPR@90\%TPR results in \Cref{app:fprtpr} to characterize the operating costs of monitoring the models at set thresholds.

% \paragraph{FPR@$X$\%TPR.}
% The False Positive Rate at a fixed True Positive Rate operating point measures the fraction of correct tool calls incorrectly flagged as erroneous when the probe is calibrated to detect $X$\% of all true errors:

% \begin{equation}
%     \mathrm{FPR}@X\%\mathrm{TPR} =
%     \frac{\mathrm{FP}(\tau^*)}{\mathrm{FP}(\tau^*) + \mathrm{TN}(\tau^*)},
% \end{equation}

% \noindent where $\tau^* = \inf\{\tau : R(\tau) \geq X/100\}$ is the lowest threshold achieving the target recall. This metric directly quantifies the operational cost of running the probe in a production setting: a lower FPR at a fixed TPR setpoint indicates fewer unnecessary interventions while maintaining a guaranteed detection rate.

\section{Experimental Setup}

\begin{figure*}[h]
    \centering
    \includegraphics[width=0.45\linewidth,trim=0 0 850 0, clip]{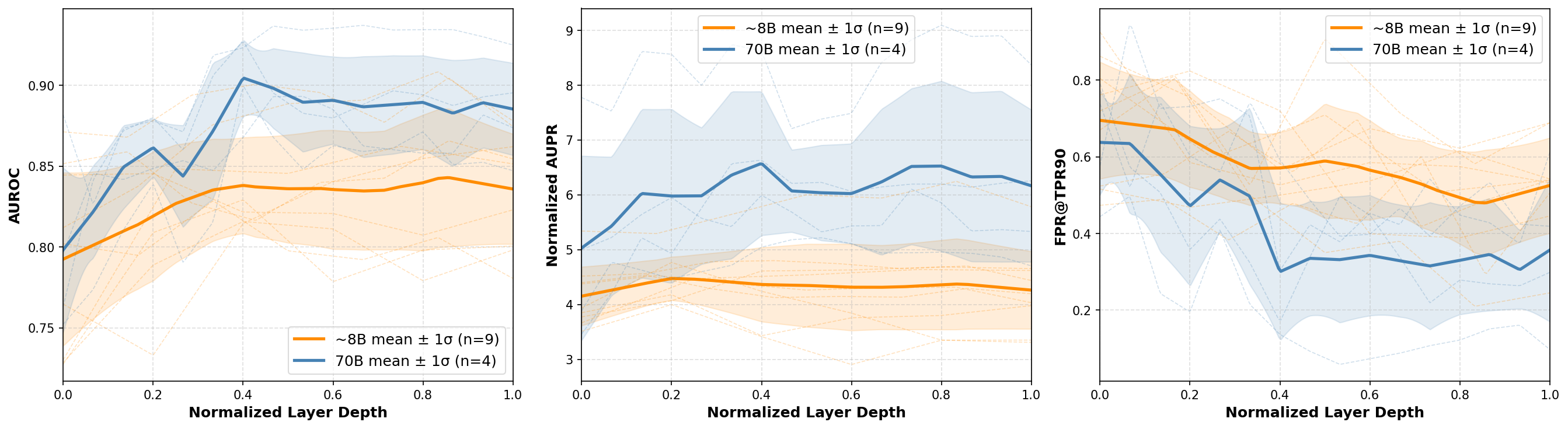}
    \hfill
    \includegraphics[width=0.52\linewidth]{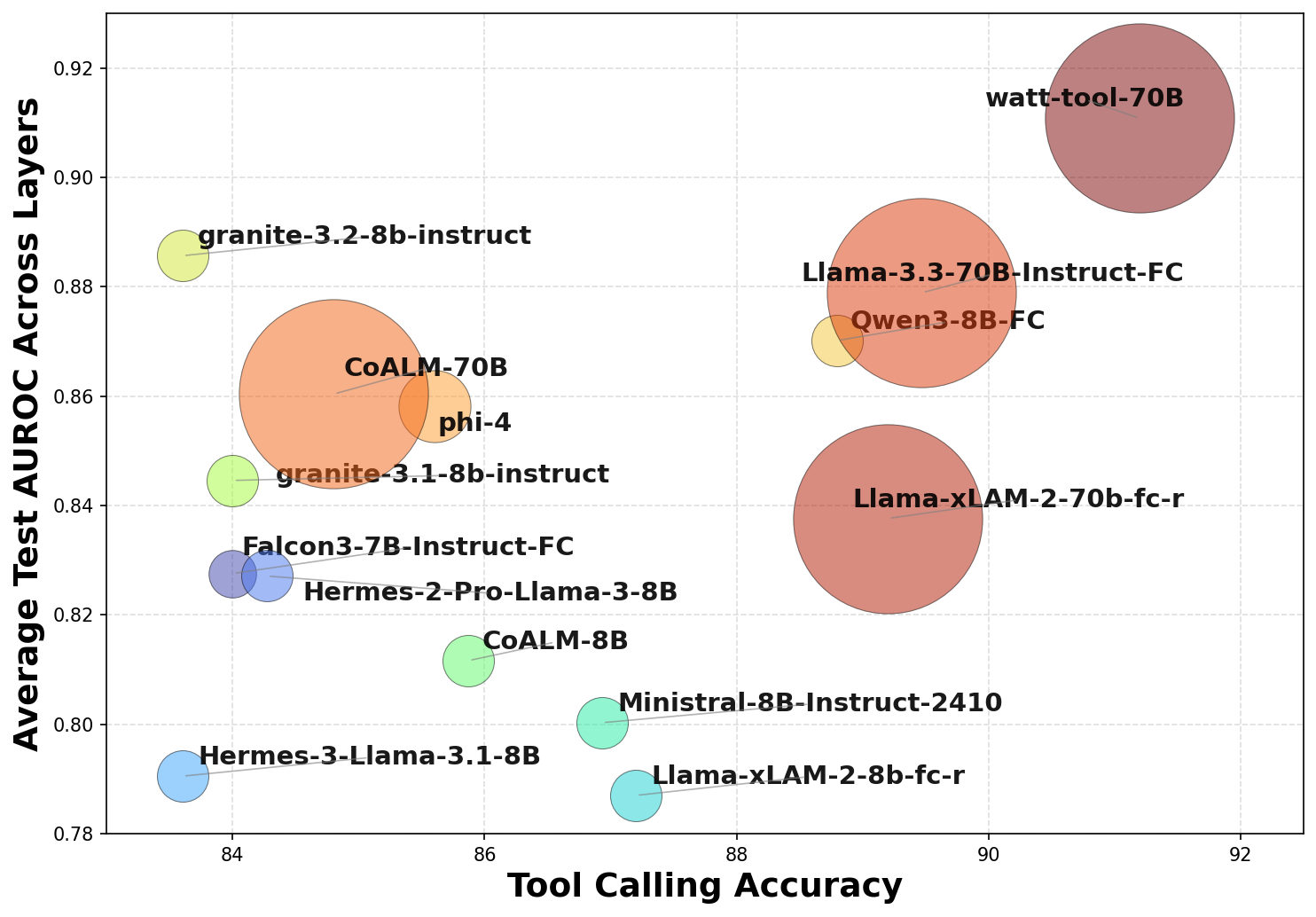}
    
    \caption {\textbf{(Left)}
 AUROC of tool-call error detection (y-axis) across hidden activation layer depth (x-axis) for smaller models ($\sim$ 8B parameters) and larger models (70B parameters). Solid line: mean. Shaded area: $\pm 1\sigma$. \textbf{(Right)} Tool-calling accuracy vs hallucination detection test AUROC (averaged across layers). Dot size corresponds to model size.}\label{fig:model_size_acc_vs_auroc_scatter}
\end{figure*}

We evaluate 18 tool-calling LLMs using the Berkeley Function Calling Leaderboard (BFCL) \citep{patil2025berkeley} on the \texttt{simple\_python}, \texttt{simple\_java}, \texttt{simple\_javascript}, and \texttt{multiple} splits (total of 750 examples per model). We save hidden states at the output of every 5th transformer layer. % We note that BFCL supplies tool definitions natively for FC-type models or via the system prompt for models that lack native FC support.

We leverage the BFCL AST checker to assign each response to one of seven outcome categories spanning a spectrum of error severity: \textbf{syntax errors} (unparseable output), \textbf{type errors} (wrong argument data type), \textbf{hallucinated tool or parameter names} (invented tool/argument not in schema), \textbf{unnecessary tool-calls} (tool invoked when none required), \textbf{missing parameter} (required argument omitted), \textbf{incorrect value} (plausible but wrong argument value), and \textbf{correct}. Some of these error types may be detected in code or prevented with constrained decoding, but error types such as incorrect value errors or unnecessary tool-calls are difficult to detect via traditional means.

For probe training and evaluation, we treat all error categories collectively as the positive class and correct tool-calls as the negative class, unless otherwise specified. We partition examples into $70\%$ train and $30\%$ test splits stratified by label to ensure balanced positive rates across splits, and we train one independent linear probe per recorded layer per model. More information is in \Cref{sec:reproducibility}.

\section{Results}\label{sec:results}

We refer the reader to \Cref{tab:tool_domain_acc} for a summary of the tool-calling performance of the 18 LLMs and \Cref{tab:error_breakdown} for a breakdown of the errors produced by each model. We exclude models with $\leq 80\%$ tool-calling accuracy (e.g., Phi-4-mini-instruct and Llama-3.1-8B-Instruct) from downstream analyses as our study is limited to models with reliable tool-calling capabilities. We plot the key metrics against \textit{normalized layer depth}, or $\ell/L$, so that models of different depths can be easily compared. Our study of the probing effectiveness vs LLM finetuning type is available in \Cref{app:finetuning_type}.

\paragraph{Effect of Model Size and Probing Layer} \Cref{fig:model_size_acc_vs_auroc_scatter} (left) shows AUROC across normalized layer depth for both smaller models ($\sim$8B parameters) and larger models ($70$B parameters). We refer the reader to \Cref{fig:aupr_results} for the corresponding normalized AUPR results. In general, most models achieve AUROC $> 0.80$ and normalized AUPR $>3.5$ across most layers, reinforcing the idea that hidden-state probes are effective at detecting tool-calling errors. Probes are substantially more effective when applied to \textit{larger models}, with average increases of $\sim 0.06$ AUROC and $\sim 2$ normalized AUPR above that of the smaller models. Probe effectiveness appears to peak in the middle to late layers of the models, suggesting that representations from the final layer (which were used in \citet{healy2026internal}) may not be optimal for tool-calling error detection. This mirrors the finding of \citet{azaria2023internal}, which studied probes in a simpler setting. More broadly, the observation that probes become more effective in a model's middle layers and as a model grows larger is consistent with the idea that probing is capturing a general representation of correctness, rather than surface-level features associated with specific tool-calling errors. The performance of the perplexity (PPL) baseline falls short of that of the probes, with a mean test AUROC of $0.62\,(\pm 0.22)$ for $\sim 8$B parameter models and a mean test AUROC of $0.80\,(\pm 0.11)$ for $70$B parameter models.

\paragraph{Tool-Calling Accuracy vs Probe Efficacy} If we assume that tool-calling ability is at least weakly correlated with model size \cite{patil2025berkeley}, one might worry that the relationship between model size and probe effectiveness is actually a by-product of the better performance of larger models. To explore this, \Cref{fig:model_size_acc_vs_auroc_scatter} (right) depicts a scatter plot of probe AUROC (averaged across layers) versus tool-calling accuracy. Each dot corresponds to a model, and the size of the dot is proportional to the number of parameters in the model. We calculate that the $R^2$ statistic within the $\sim$8B model group (for which there are 9 models) is $0.005$, suggesting little apparent correlation. On the other hand, the larger CoALM-70B and phi-4 (14B parameters) models have relatively low tool-calling accuracy of about $85\%$, but the AUROC scores of their probes are higher by approximately $+0.08$. Taken together, these results suggest that tool-calling accuracy alone is not the cause of higher error probing effectiveness.

\begin{figure}[t]
    \centering
    \includegraphics[width=0.49\linewidth,trim=5 0 850 0, clip]{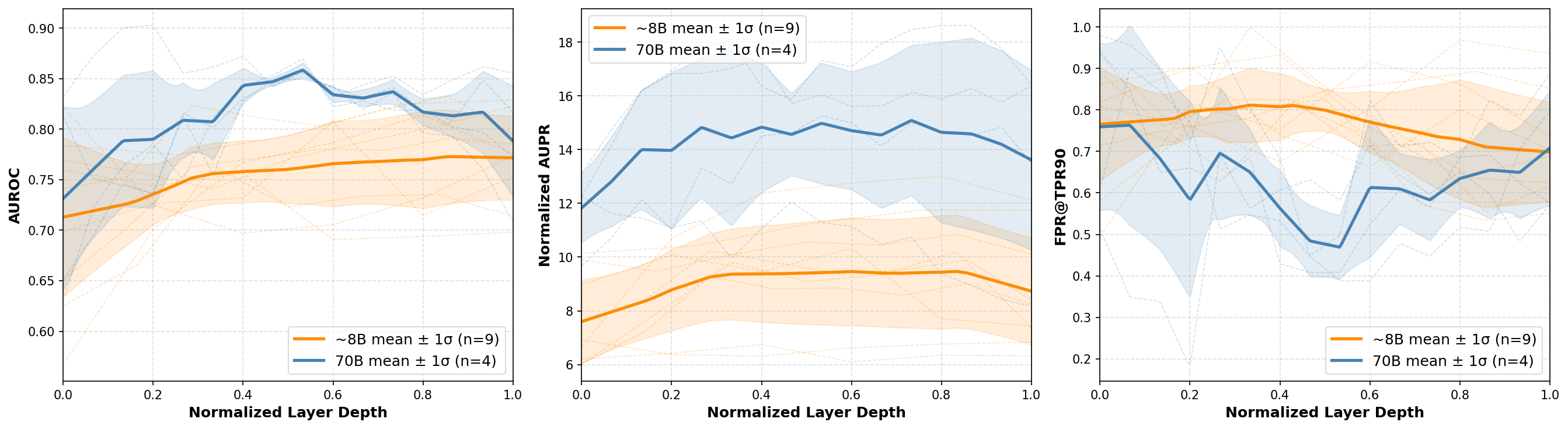}
    \hfill
    \includegraphics[width=0.49\linewidth,trim=5 0 850 0, clip]{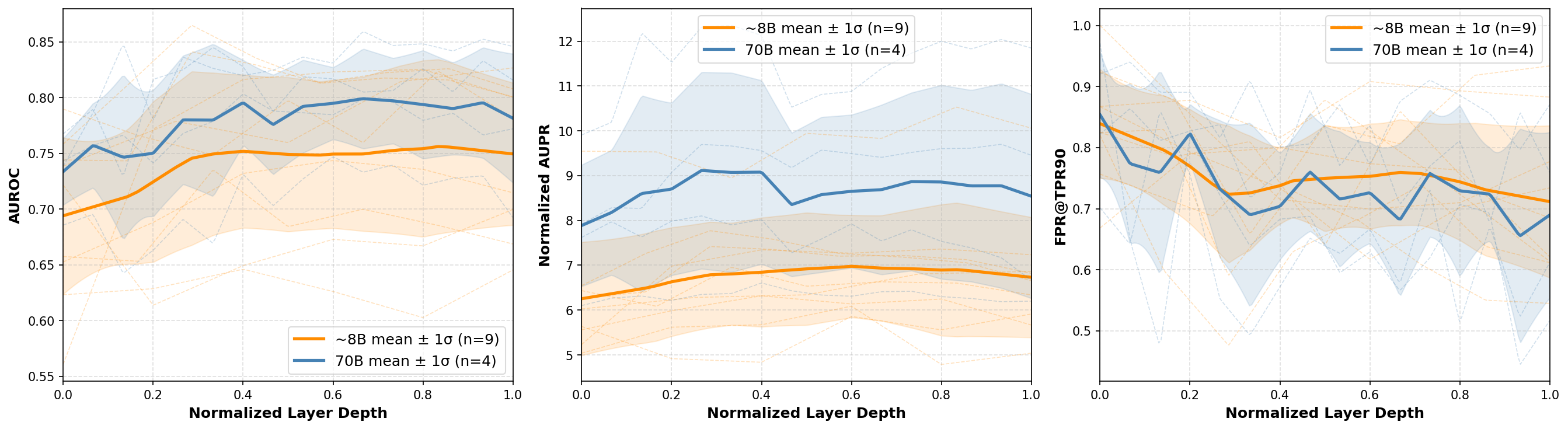}
    \caption {Label space transfer experiment. \textbf{Left}: Test AUROC for training on split A and evaluating on split B. \textbf{Right}: Test AUROC for training on split B and evaluating on split A.}\label{fig:transfer_ab}
\end{figure}

\paragraph{Behavior of Probes on Novel Error Types} Because not every failure type can be anticipated before deployment, it is critical that techniques for monitoring LLMs generalize to new behaviors. To explore this in the context of probing for tool-calling errors, we conduct a controlled experiment in which we train probes on one subset of error types (e.g., incorrect value errors and missing parameter errors) and evaluate them on a disjoint subset of error types (e.g., type errors and hallucinated tools/parameters). \Cref{fig:transfer_ab} depicts the test AUROC of probes for large models and smaller models in this experiment. The left plot depicts the metrics for probes trained on ``Split A'' and evaluated on ``Split B'', and the right plot depicts the metrics for probes trained on ``Split B'' and evaluated on ``Split A''. We refer the reader to \Cref{fig:aupr_transfer_ab} for the corresponding normalized AUPR scores. For information on the makeup of the splits and results on additional ``C'' and ``D'' splits, we refer the reader to \Cref{app:cd_label_transfer}.

\Cref{fig:transfer_ab} shows AUROC scores greater than 0.75, and \Cref{fig:aupr_transfer_ab} shows normalized AUPR scores greater than 6 for many of the models in both transfer of A to B and B to A, suggesting that these probes are learning features associated with general tool-calling correctness rather than specific types of tool-calling errors. Additionally, probes on large models achieve significantly better transfer with AUROC increases of $\sim 0.05$-$0.1$ and normalized AUPR increases of $10$-$40\%$ relative to smaller models. Middle layers of large models appear to achieve slightly better transfer performance. This aligns with the findings of \citet{skean2025layer}, that intermediate layers of LLMs encode rich representations useful on a range of downstream tasks. Again, the performance of the perplexity (PPL) baseline falls short of that of the probes. For the A$\rightarrow$B transfer task, PPL yields a mean test AUROC of $0.61\,(\pm 0.21)$ for $\sim 8$B parameter models and a mean test AUROC of $0.76\,(\pm 0.05)$ for $70$B parameter models. For the B$\rightarrow$A transfer task, PPL yields a mean test AUROC of $0.63\,(\pm 0.23)$ for $\sim 8$B parameter models and a mean test AUROC of $0.80\,(\pm 0.08)$ for $70$B parameter models.

\section{Conclusion}

In this paper, we explore the use of linear probes for the detection of tool-calling errors. Our experiments indicate that model size, probing layer, and post-training regime are associated with the effectiveness of probes. Significantly, our results suggest that probes from middle layers capture features associated with abstract representations of tool-calling correctness, allowing them to generalize to error types unseen during training. These findings provide further evidence that probes are an essential component of a robust monitoring strategy for LLMs. We hope this work encourages further exploration of how hidden activations can be leveraged for more reliable agentic systems.

\bibliographystyle{plainnat}
\bibliography{custom}

%%%%%%%%%%%%%%%%%%%%%%%%%%%%%%%%%%%%%%%%%%%%%%%%%%%%%%%%%%%%

\appendix

\section{Limitations}

Our exploratory study is limited in dataset scope and scale. The BFCL splits used cover relatively clean, well-defined function calling schemas, which may not reflect the complexity of real production tool-calls involving longer contexts, multi-turn conversations, or ambiguous schemas. With 750 examples per model, some error categories have few positive examples, making AUPR estimates potentially noisy for rare error types. Additionally, all models studied are open-weight, and the findings may not generalize to closed models.

Our probe design choices are not fully ablated. We study only linear probes and extract hidden states exclusively at the last token position, following prior work, but do not compare against nonlinear probes or alternative token aggregation strategies. It is possible that these choices interact with model size or finetuning type in ways not captured by our analysis.

Finally, our findings are correlational rather than causal. Models differ along many axes beyond size and finetuning type, including base model, training data composition, and alignment procedure, which are confounded in our cross-model comparisons.

\section{Metric Background} \label{sec:metrics}

\paragraph{AUROC.} The Area Under the Receiver Operating Characteristic curve (AUROC) measures the probability that a randomly chosen incorrect tool-call is assigned a higher score by the probe than a randomly chosen correct one:
\begin{equation}
    \mathrm{AUROC} = \mathbb{P}\!\left(
        p_\theta(y{=}1 \mid \mathbf{z}^+) >
        p_\theta(y{=}1 \mid \mathbf{z}^-)
    \right),
\end{equation}
\noindent where $\mathbf{z}^+$ and $\mathbf{z}^-$ denote hidden states from incorrect and correct tool-calls, respectively. AUROC ranges from 0 to 1, with 0.5 indicating random performance and 1.0 indicating perfect discrimination. Crucially, AUROC is invariant to class imbalance, making it a reliable metric for cross-model comparisons where the base rate of errors varies.

\paragraph{AUPR.}
The Area Under the Precision-Recall curve (AUPR) summarizes the tradeoff between precision $P(\tau)$ and recall $R(\tau)$ across all decision thresholds $\tau$:
\begin{equation}
    \mathrm{AUPR} = \int_0^1 P(\tau_R)\, dR,
\end{equation}
\noindent where $\tau_R$ is the decision threshold corresponding to a set recall level. Unlike AUROC, the random baseline for AUPR equals the base positive rate $\rho = |\{i : y_i = 1\}| / N$. We normalize AUPR with $\rho$ to provide a comparison across models. One can interpret this \textit{normalized} AUPR as the relative advantage of a detector over the random guessing baseline.

\section{Additional Reproducibility Information}\label{sec:reproducibility}

We evaluate 18 tool-calling LLMs using the Berkeley Function Calling Leaderboard (BFCL) \citep{patil2025berkeley} on the \texttt{simple\_python} (400 examples), \texttt{simple\_java} (100 examples), \texttt{simple\_javascript} (50 examples), and \texttt{multiple} (200 examples) splits (750 examples per model total).

\paragraph{35k Additional Glaive Data} We also filter and extract user prompts and gold-standard python tool calling instances from the ``\texttt{glaiveai/glaive-function-calling-v2}'' \href{https://huggingface.co/datasets/glaiveai/glaive-function-calling-v2}{dataset on HuggingFace}. The original 113k prompts and responses were filtered based on the following criteria: (1) the assistant's first response must contain one function call, and (2) the function call must take arguments. This ensures that the filtered glaive data match our single-turn BFCL format and that the gold-standard function calls are not trivial. We then translate the filtered function calls into BFCL's format. This resulted in 35,781 prompt and response pairs from the glaive data. Any subsequent chat turns were discarded, simulating single-turn function calling data. Following BFCL's naming style, we call this domain ``\texttt{glaive\_simple\_python}''.

\paragraph{Hidden States Collection} To accommodate the large number of models and hidden states that we record in this study, we separate response generation and hidden states extraction into a two stage process. First, we leverage vLLM (already integrated in BFCL \cite{patil2025berkeley}) to rapidly generate the model tool-calls. Unfortunately, vLLM does not provide access to the hidden states of all layers - at the time of this research, vLLM only provides the hidden states of the last layer. To collect the hidden states of other layers, we use teacher-forcing of the prior prompt and response (with each model's specific chat template) for a version of the model implemented in HuggingFace \texttt{transformers}. This is a rapid process, since the \texttt{transformers} model merely encodes representations of the prompt and response and no autoregressive generation is performed in the slow \texttt{transformers} implementation. This allows us to collect the prompts, responses, and hidden states from each model quickly.

Additionally, to reduce storage requirements, we collect just the hidden states of the final token of every 5th layer of each model. Hence, we collect the hidden states from layers 5, 10, 15, 20, 25, etc. 

\paragraph{Probe Information} We employ \texttt{LogisticRegression} from \texttt{sklearn} with standard normalization, $\ell_2$ regularization parameter $C=1$, and 2000 iterations with the \texttt{``liblinear''} solver when training probes. Probes are evaluated on test data using the normalization fit in the training phase. 

\section{Tool-call Performance Results}\label{app:tool_call_performance}

\begin{figure*}[h]
    \centering
    \includegraphics[width=0.95\linewidth,trim=0 0 425 0, clip]{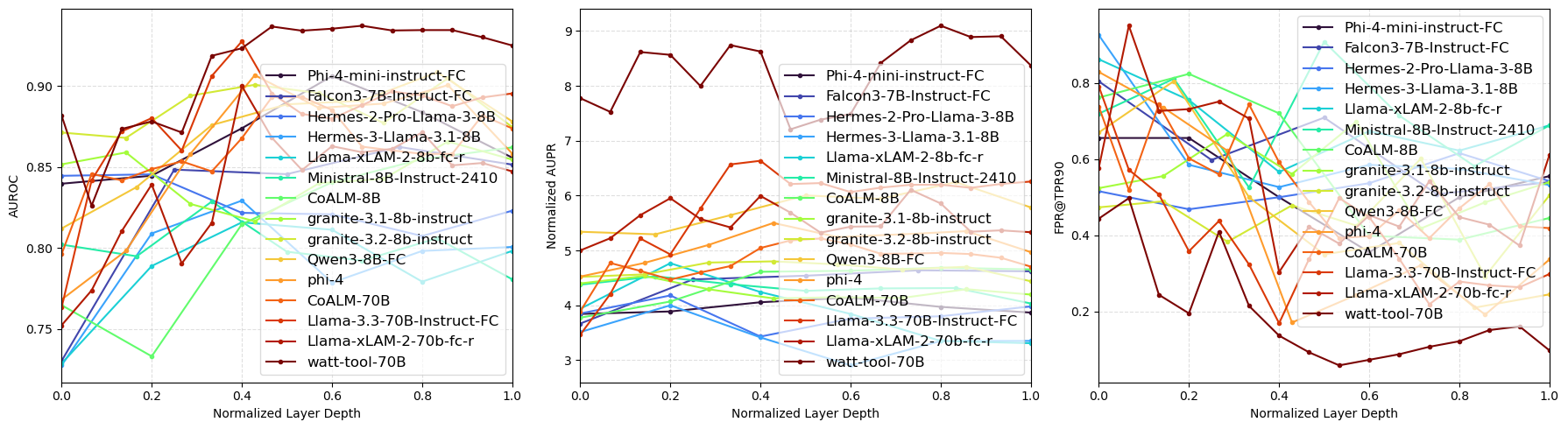}
    \caption {Probe tool-calling error detection performance for all models (colored by model size).}
\end{figure*}

\begin{table*}[h]
  \centering
  \small
  \begin{tabular}{llrrrr}
    \hline
    \textbf{Model} & \textbf{Size} & \textbf{Python Tools} & \textbf{Java Tools} & \textbf{JavaScript Tools} & \textbf{Python Tools} \\
                   & \textbf{(B)}  & \textbf{(Single)}     & \textbf{(Single)}   & \textbf{(Single)}         & \textbf{(Multi)} \\
    \hline
    \textcolor{gray}{gemma-3-1b-it}          & 1   & 24.00 & 5  & 4  & 0.00  \\
    \textcolor{gray}{Phi-4-mini-instruct}    & 3.8 & 84.75 & 54 & 62 & 88.50 \\
    \hline
    Falcon3-7B-Instruct    & 7   & 93.25 & 58 & 42 & 89.00 \\
    \textcolor{gray}{Llama-3.1-8B-Instruct}  & 8   & 50.50 & 57 & 60 & 54.50 \\
    granite-3.1-8b         & 8   & 90.50 & 57 & 56 & 91.50 \\
    granite-3.2-8b         & 8   & 91.75 & 54 & 60 & 88.00 \\
    Ministral-8B-2410      & 8   & 93.75 & 60 & 66 & 92.00 \\
    Hermes-2-Pro-Llama-3   & 8   & 90.00 & 61 & 56 & 91.50 \\
    Hermes-3-Llama-3.1     & 8   & 88.75 & 59 & 56 & 92.50 \\
    CoALM-8B               & 8   & 92.50 & 60 & 54 & 93.50 \\
    xLAM-2-8b-fc-r         & 8   & 92.50 & 65 & 64 & 93.50 \\
    Qwen-3-8B              & 8   & 96.00 & 60 & 62 & 95.50 \\
    \hline
    phi-4                  & 14  & 93.50 & 53 & 74 & 89.00 \\
    \textcolor{gray}{granite-20b-code}       & 20  & 37.75 & \textbf{66} & 64 & 91.50 \\
    \hline
    CoALM-70B              & 70  & 89.50 & 64 & 58 & 92.50 \\
    xLAM-2-70b-fc-r        & 70  & 94.75 & 64 & \textbf{76} & 94.00 \\
    Llama-3.3-70B-Instruct & 70  & 96.25 & 58 & \textbf{76} & 95.00 \\
    watt-tool-70B          & 70  & \textbf{97.50} & 65 & 74 & \textbf{96.00} \\
    \hline
  \end{tabular}
  \caption{\label{tab:tool_domain_acc}
    Tool-call accuracy (\%) by model and evaluation setting. Models are sorted by size.
    \textit{Single} denotes single tool use; \textit{Multi} denotes selecting and correctly employing one tool from multiple candidates. Provider prefixes are omitted from model names for brevity. Models with high error rates ($\geq 20\%$ of all evaluation samples) are excluded from probe-based analysis and have their names grayed out.
  }
\end{table*}

\Cref{tab:tool_domain_acc} depicts the domain-level accuracy of all tool-calling LLMs in our study. The models are organized by parameter count, and models with \textcolor{gray}{grayed-out names} are omitted from our probe analyses due to having excessive ($\geq 20\%$) error rates.

\begin{table*}[h]
  \centering
  \small
  \begin{tabular}{llrrrrrrrr}
    \hline
    \textbf{Model} & \textbf{Size} & \textbf{Correct} & \textbf{Incor.} & \textbf{Missing} & \textbf{Unnec.} & \textbf{H. Tool/} & \textbf{Type} & \textbf{Syntax} \\
                   & \textbf{(B)}  &                  & \textbf{Value}     & \textbf{Param.}  & \textbf{Tool}   & \textbf{Param.}  & \textbf{Error} & \textbf{Error} \\
    \hline
    \textcolor{gray}{gemma-3-1b-it}         & 1   & 13.73 & 3.74 & 22.00 & 0.26 & 6.27 & 1.86  & 52.14 \\
    \textcolor{gray}{Phi-4-mini-instruct}   & 3.8 & 80.13 & 5.07 & 0.40  & 0.40 & 0.53 & 3.33  & 10.13 \\
    \hline
    Falcon3-7B-Instruct   & 7   & 84.00 & 9.46 & 0.80  & 0.94 & 0.27 & 4.14  & 0.40  \\
    \textcolor{gray}{Llama-3.1-8B-Instruct} & 8   & 53.07 & 7.20 & 0.00  & 0.27 & 0.80 & 37.07 & 1.60  \\
    granite-3.1-8b        & 8   & 84.00 & 6.13 & 0.66  & 0.67 & 0.40 & 5.20  & 2.93  \\
    granite-3.2-8b        & 8   & 83.60 & 5.99 & 0.66  & 0.80 & 0.27 & 5.33  & 3.33  \\
    Ministral-8B-2410     & 8   & 86.93 & 4.80 & 1.60  & 0.27 & 0.26 & 4.40  & 1.73  \\
    Hermes-2-Pro-Llama-3  & 8   & 84.27 & 5.73 & 1.20  & 3.73 & 0.27 & 4.80  & 0.00  \\
    Hermes-3-Llama-3.1    & 8   & 83.60 & 5.87 & 1.74  & 1.60 & 2.13 & 5.07  & 0.00  \\
    CoALM-8B              & 8   & 85.87 & 6.40 & 1.07  & 0.27 & 0.40 & 3.33  & 2.67  \\
    xLAM-2-8b-fc-r        & 8   & 87.20 & 5.46 & 1.47  & 0.40 & 0.26 & 4.53  & 0.67  \\
    Qwen3-8B              & 8   & 88.80 & 5.46 & 0.13  & 0.93 & 0.40 & 4.13  & 0.13  \\
    \hline
    Phi-4                 & 14  & 85.60 & 3.19 & 0.27  & 0.53 & 1.06 & 2.26  & 7.07  \\
    \textcolor{gray}{granite-20b-code}      & 20  & 57.60 & 3.34 & 0.53  & 1.33 & 0.40 & 5.07  & 31.73 \\
    \hline
    CoALM-70B             & 70  & 84.80 & 5.86 & 0.80  & 0.40 & 0.27 & 3.07  & 4.80  \\
    xLAM-2-70b-fc-r       & 70  & 89.20 & 6.00 & 0.13  & 0.13 & 0.13 & 4.00  & 0.40  \\
    Llama-3.3-70B-Instruct& 70  & 89.47 & 4.79 & 0.00  & 0.67 & 0.40 & 2.66  & 2.00  \\
    watt-tool-70B         & 70  & 91.20 & 4.12 & 0.27  & 0.26 & 0.27 & 2.80  & 1.07  \\
    \hline
  \end{tabular}
  \caption{\label{tab:error_breakdown}
    Tool-call error breakdown by model (in \%). Models are sorted by size.
    \textit{Correct} indicates fully correct tool-calls; all other columns indicate error categories. Provider prefixes are omitted from model names for brevity. Models with high error rates ($\geq 20\%$ overall) are excluded from probe-based analysis and have their names grayed out.
  }
\end{table*}

\Cref{tab:error_breakdown} depicts the breakdown of error types for each model. The error breakdown is computed across all tool-calling domains (e.g., \texttt{simple\_python}, \texttt{simple\_javascript}, etc). The prevalence of errors of different type can vary somewhat between models, but most models consistently generate incorrect value errors and type errors.

\begin{table*}[h]
  \centering
  \small
  \begin{tabular}{llrrrrrrr}
    \hline
    \textbf{Model} & \textbf{Size} & \textbf{Correct} & \textbf{Incor.} & \textbf{Missing} & \textbf{Unnec.} & \textbf{H. Tool/} & \textbf{Type} & \textbf{Syntax} \\
                   & \textbf{(B)}  &                  & \textbf{Value}  & \textbf{Param.}  & \textbf{Tool}   & \textbf{Param.}   & \textbf{Error} & \textbf{Error} \\
    \hline
    \textcolor{gray}{Llama-3.1-8B-Instruct} & 8 & 71.79 & 6.21 & 0.16 & 0.01 & 2.10 & 19.14 & 0.60 \\
    xLAM-2-8b-fc-r                      & 8 & 94.95 & 2.89 & 0.86 & 0.00 & 0.41 & 0.04  & 0.85 \\
    Hermes-2-Pro-Llama-3                & 8 & 94.53 & 3.53 & 0.42 & 0.99 & 0.48 & 0.06  & 0.00 \\
    Hermes-3-Llama-3.1                    & 8 & 94.94 & 2.80 & 0.26 & 1.04 & 0.91 & 0.06  & 0.00 \\
    Qwen3-8B                       & 8 & 93.38 & 4.77 & 0.53 & 0.79 & 0.49 & 0.03  & 0.00 \\
    \hline
  \end{tabular}
  \caption{\label{tab:error_breakdown_glaive}
    Tool-call error breakdown by model on \texttt{glaive\_simple\_python} (in \%).
    \textit{Correct} indicates fully correct tool-calls; all other columns indicate error categories. Provider prefixes are omitted from model names for brevity.
  }
\end{table*}

\Cref{tab:error_breakdown_glaive} depicts the breakdown of error types for each model on the \texttt{glaive\_simple\_python} set. The prevalence of errors of different type can vary somewhat between models, but most models consistently generate incorrect value errors. One exception is Llama-3.1-8B-Instruct, which generates a large number of type errors.

\newpage

\section{Effect of Model Finetuning on Probe Efficacy}\label{app:finetuning_type}

\begin{table}[h]
  \centering
  \small
  \begin{tabular}{ll}
    \hline
    \textbf{Model} & \textbf{Post-Training} \\
    \hline
    Falcon3-7B-Instruct    & Instruction Tuned \\
    granite-3.1-8b         & Instruction Tuned \\
    granite-3.2-8b         & Instruction Tuned \\
    Ministral-8B-2410      & Instruction Tuned \\
    Qwen3-8B               & Instruction Tuned \\
    \hline
    Hermes-2-Pro-Llama-3   & Tool Finetuned \\
    Hermes-3-Llama-3.1     & Tool Finetuned \\
    xLAM-2-8b-fc-r         & Tool Finetuned \\
    CoALM-8B               & Tool Finetuned \\
    \hline
  \end{tabular}
  \caption{\label{tab:finetuning_type}
    Classification of $\sim$8B parameter models by post-training regime.
    Tool finetuned models have been explicitly trained on function calling
    or agent action datasets. Provider prefixes are omitted for brevity.
  }
\end{table}

\begin{figure}
    \centering
    \includegraphics[width=0.9\linewidth,trim=0 0 425 0, clip]{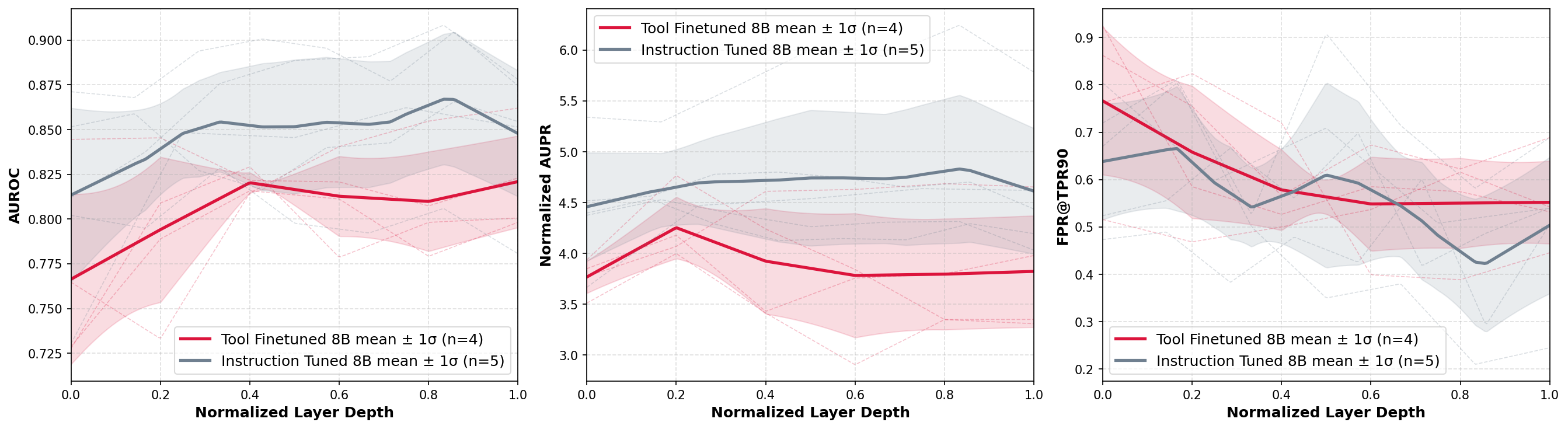}
    \caption{Test AUROC (left) and Normalized AUPR (right) of tool-call error detection across recording layer depth for instruction tuned vs tool finetuned models (all $\sim8$B parameter models). Solid line: mean. Shaded area: $\pm 1\sigma$.}\label{fig:model_size_finetuning}

\end{figure}

\begin{figure*}[h]
    \centering
    \includegraphics[width=0.7\linewidth,trim=435 0 430 0, clip]{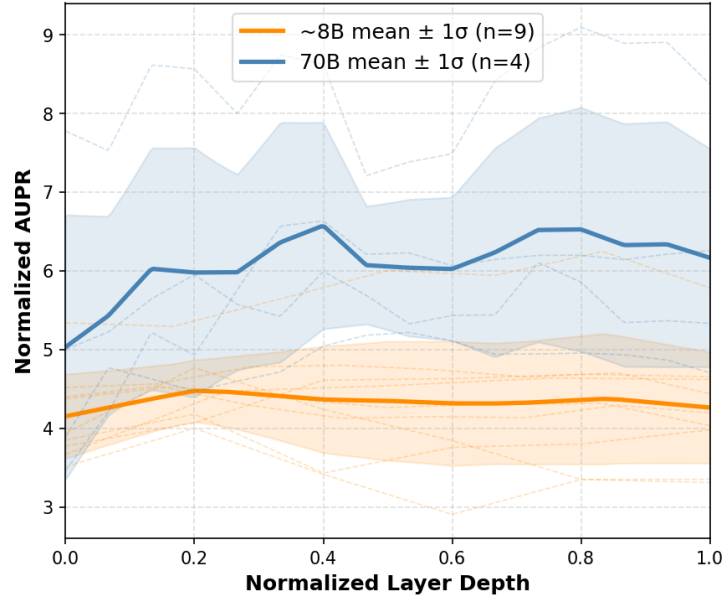}
    
    \caption {Normalized AUPR of tool-call error detection (y-axis) across hidden activation layer depth (x-axis) for smaller models ($\sim$ 8B parameters) and larger models (70B parameters). Solid line: mean. Shaded area: $\pm 1\sigma$.}\label{fig:aupr_results}
\end{figure*}

We divide the $\sim$8B parameter models into those finetuned specifically for tool-calling tasks and those finetuned for general instruction following (see \Cref{tab:finetuning_type} for details on the models considered). We then compare the effectiveness of probes (measured by AUROC) in detecting tool-calling errors between these two groups (\Cref{fig:model_size_finetuning}). We refer the reader to \Cref{fig:aupr_results} for the corresponding normalized AUPR results. Surprisingly, we find that probes trained and applied to general instruction-tuned models (gray) are slightly more effective than probes trained and applied to tool-finetuned models (red), with an AUROC increase of $+0.05$ and normalized AUPR increase of $+0.5$ respectively. We emphasize that these findings are correlational, and other factors such as the volume of training data or choice of training methodology could also influence the probing results.

We conducted a controlled experiment comparing Llama-3.1-8B-Instruct (general instruction-tuned) against Hermes-3-Llama-3.1-8B (tool-finetuned from the same base model). The results are consistent with the trend in the correlational study, with the general instruction-tuned model yielding substantially more effective probes. Probes on Llama-3.1-8B-Instruct yielded AUROCs of 90-95\% and an FPR@90\%TPR of 15-35\% while probes on Hermes-3-Llama-3.1-8B yielded AUROCs of 72-83\% and FPR@90\%TPR of 50-60\%. 

\section{Label Space Transfer Experiments}\label{app:cd_label_transfer}

The label space transfer experiments partition the seven outcome categories into two disjoint subsets, which we refer to as splits. For the A/B partition, Split A consists of incorrect value, missing parameter, and syntax errors, while Split B consists of unnecessary tool-call, hallucinated tool or parameter, and type errors. For the C/D partition, Split C consists of missing parameter, type error, and syntax error, while Split D consists of incorrect value, unnecessary tool-call, and hallucinated tool or parameter. The splits were designed so that the aggregate positive rate (i.e., the fraction of incorrect tool-calls) is approximately balanced between the two disjoint subsets for most models in the study, enabling a fair comparison of probe transfer in both directions. Note that the correct category is always treated as the negative class, and its examples are evenly split between training and testing.

\begin{figure*}[h]
    \centering
    \includegraphics[width=0.95\linewidth,trim=0 0 425 0, clip]{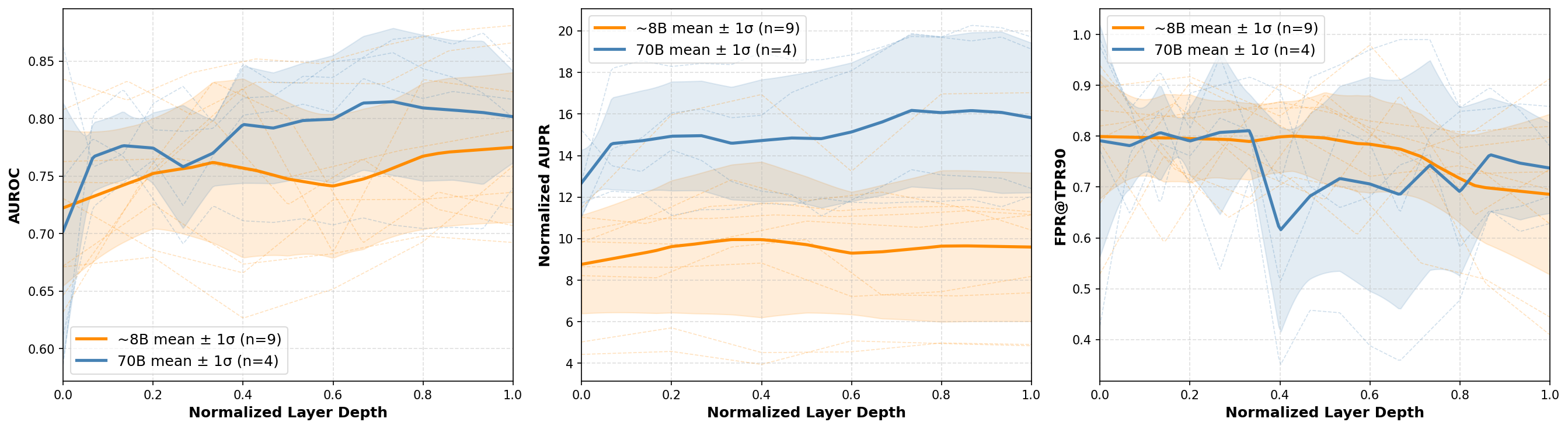}
    
    \includegraphics[width=0.95\linewidth,trim=0 0 425 0, clip]{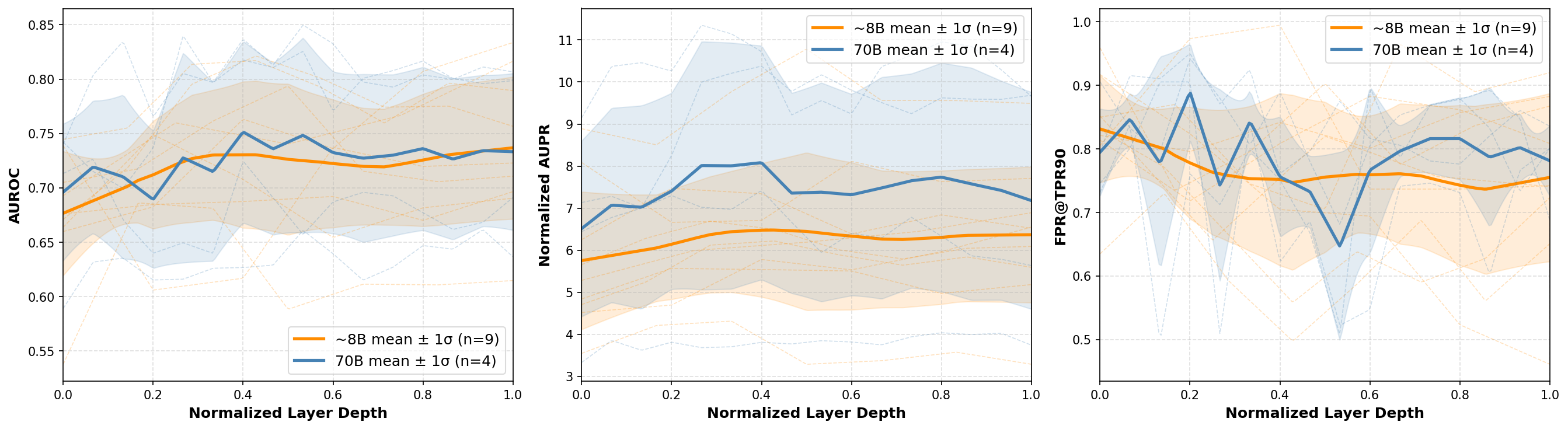}
    \caption {Label space transfer experiment. Top row: metrics for C to D. Bottom row: metrics for D to C. Solid line: mean. Shaded area: $\pm 1\sigma$.}\label{fig:transfer_cd}
\end{figure*}

\Cref{fig:transfer_cd} depicts the probe efficacy metrics (AUROC and normalized AUPR) for the C/D partition of the tool-calling data. The trends are similar: probes on larger models appear as good or better at separating the novel errors from correct tool-calls compared with probes on smaller models.

\begin{table*}[h]
  \centering
  \small
  \begin{tabular}{llp{10cm}}
    \hline
    \textbf{Partition} & \textbf{Split} & \textbf{Error Categories} \\
    \hline
    A/B & A & Incorrect Value, Missing Parameter, Syntax Error \\
        & B & Unnecessary Tool-Call, Hallucinated Tool/Param., Type Error \\
    \hline
    C/D & C & Missing Parameter, Type Error, Syntax Error \\
        & D & Incorrect Value, Unnecessary Tool-Call, Hallucinated Tool/Param. \\
    \hline
  \end{tabular}
  \caption{\label{tab:splits}
    Composition of label space transfer splits. Each split contains
    three of the six error categories; the correct category is always
    treated as the negative class. Within each partition, splits are
    designed to approximately balance the aggregate positive rate
    across the two subsets for most models.
  }
\end{table*}

\begin{figure*}[h]
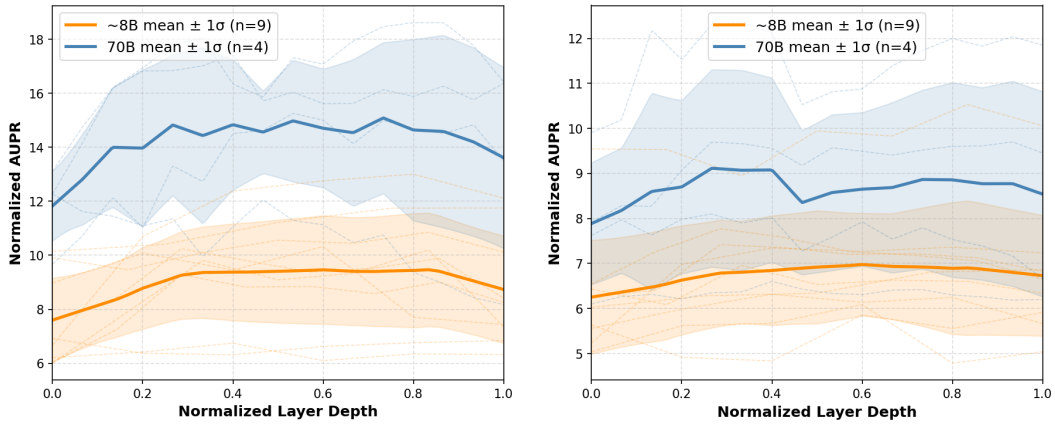

    \centering
    \includegraphics[width=0.49\linewidth,trim=435 0 430 0, clip]{figs/transfer_probe_metrics_by_layer_grouped_ab.png}
    \hfill
    \includegraphics[width=0.49\linewidth,trim=435 0 430 0, clip]{figs/transfer_probe_metrics_by_layer_grouped_ba.png}
    \caption {Normalized AUPR for the A/B label space transfer experiment. Left: normalized AUPR for A to B. Right: normalized AUPR for B to A. Solid line: mean. Shaded area: $\pm 1\sigma$.}\label{fig:aupr_transfer_ab}
\end{figure*}

\section{Glaive Probe Performance Experiments}\label{app:glaive_experiments}

\begin{figure*}[h]

    \centering
    \begin{subfigure}{0.49\textwidth}
        \includegraphics[width=\textwidth]{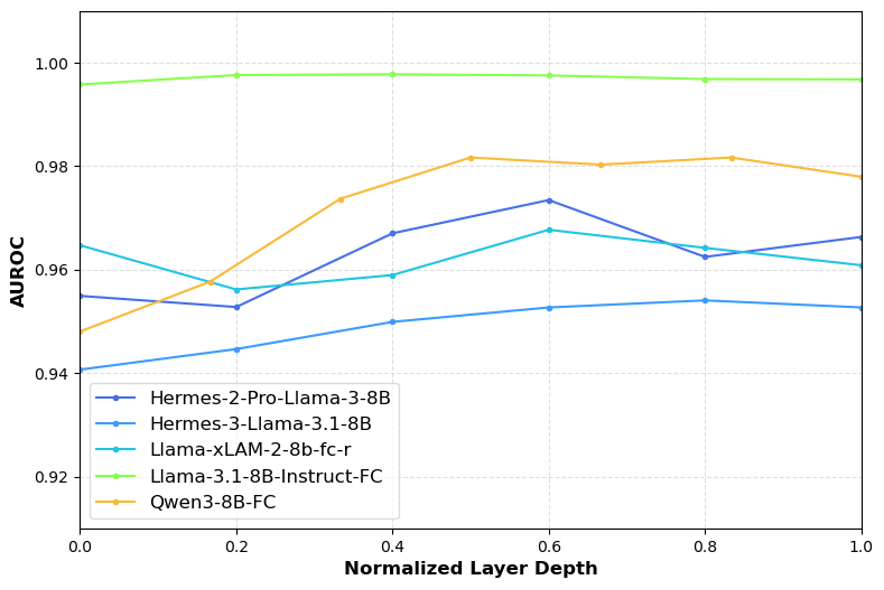}
        \caption{Probe Test AUROC on all error types vs. normalized probing layer depth.}
    \end{subfigure}
    \hfill
    \begin{subfigure}{0.49\textwidth}
        \includegraphics[width=\textwidth]{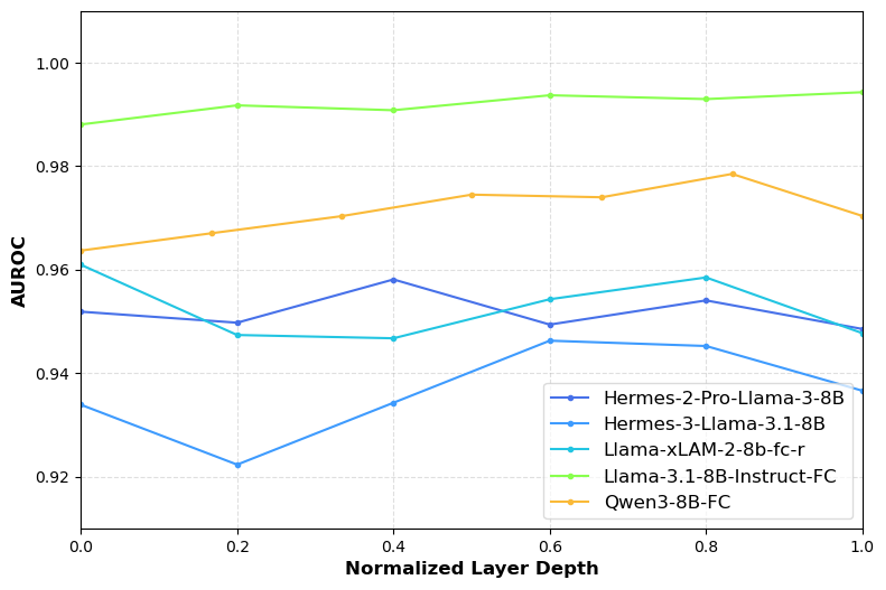}
        \caption{Probe Test AUROC on incorrect value errors vs. normalized probing layer depth.}
    \end{subfigure}
    \caption{Glaive probe performance experiment for all error types (left) and only incorrect value errors (right). Probes are slightly more effective on all error types, indicating the incorrect value errors are typically more challenging to detect.}\label{fig:glaive_probe_results}

\end{figure*}

\Cref{fig:glaive_probe_results} reports the test AUROC of probes operating at various normalized layer depths for five of the 8B parameter models. The figure on the left reports test AUROC for probes trained and evaluated on all error types while the figure on the right reports test AUROC for probes trained and evaluated on just the incorrect value error type. In both settings, the test AUROC is high (0.92-0.98), indicating that probes are effective at detecting various error types on the Glaive set. Test AUROC for probes tends to be slighly better in the multi-error setting (left) compared with the incorrect value error setting (right). This indicates that incorrect value errors are more difficult for the probes to distinguish from correct answers compared with all error types.

\begin{figure*}[h]

    \centering
    \includegraphics[width=0.9\textwidth]{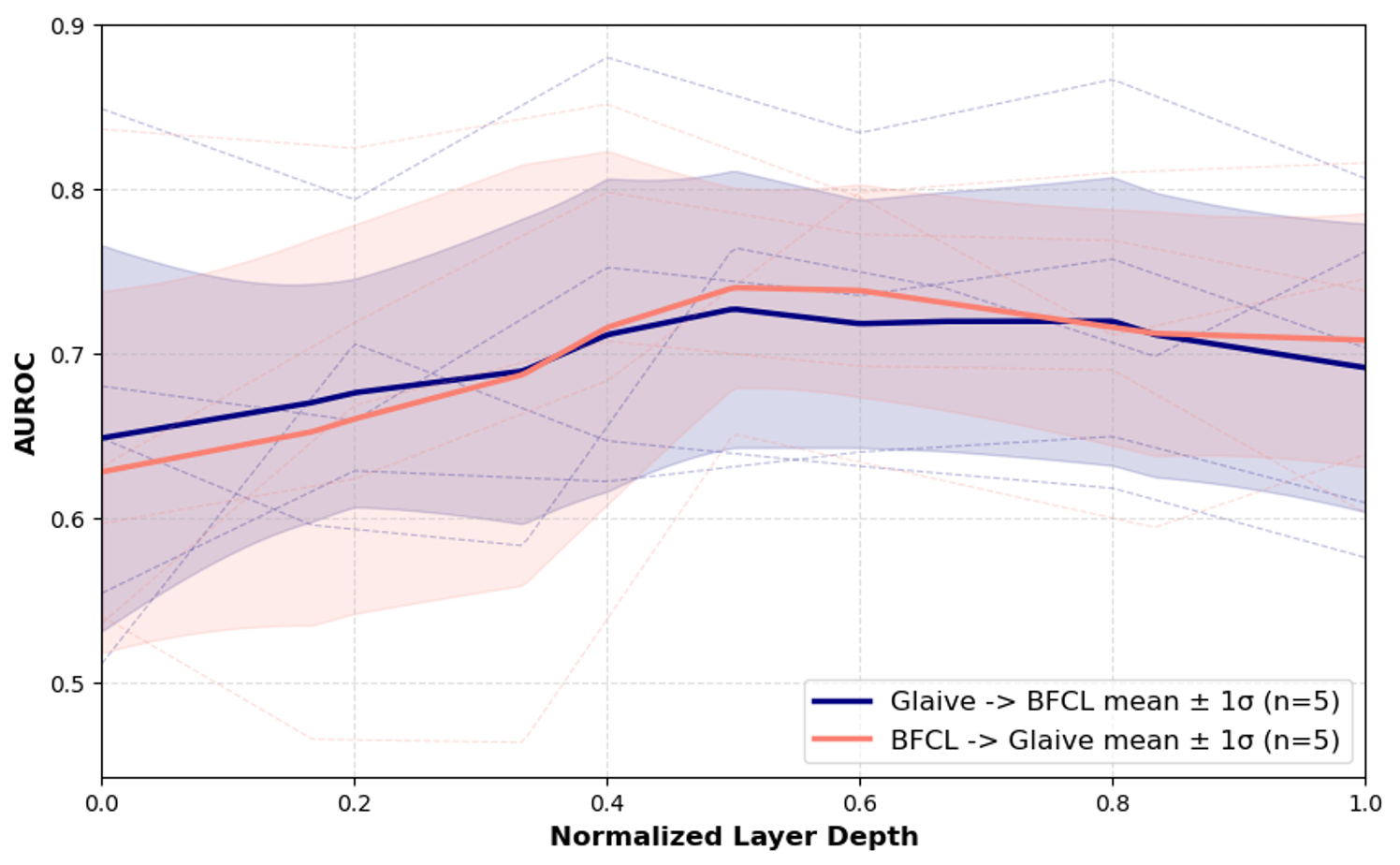}
    \caption{Glaive to BFCL and BFCL to Glaive probe transfer performance experiment. The probes are trained to distinguish incorrect value tool-calls from correct tool-calls. Test AUROC (y-axis) is reported for probes deployed at various normalized layer depths (x-axis). Solid line: mean. Shaded area: $\pm 1\sigma$. }\label{fig:glaive_probe_transfer_results}

\end{figure*}

\Cref{fig:glaive_probe_transfer_results} reports the cross-domain test AUROC of incorrect value probes trained on either Glaive or BFCL. The purple line and shaded area represent probes trained on Glaive data and evaluated on BFCL data. The orange line and shaded area represent probes trained on BFCL data and evaluated on Glaive data. Cross-domain probe transfer performance peaks at $\sim 70-75\%$ at a normalized layer depth of $50-60\%$. This is a 10 percentage point increase in test AUROC compared with deploying probes in the initial layers, and it agrees with prior observations that more abstract, transferable processing occurs in central layers \cite{skean2025layer}.

\newpage
\section{Results on False Positive Rate (FPR) at 90\% True Positive Rate (TPR)}\label{app:fprtpr}

\begin{figure}[h]
    \centering
    \includegraphics[width=0.8\linewidth,trim=855 0 0 0, clip]{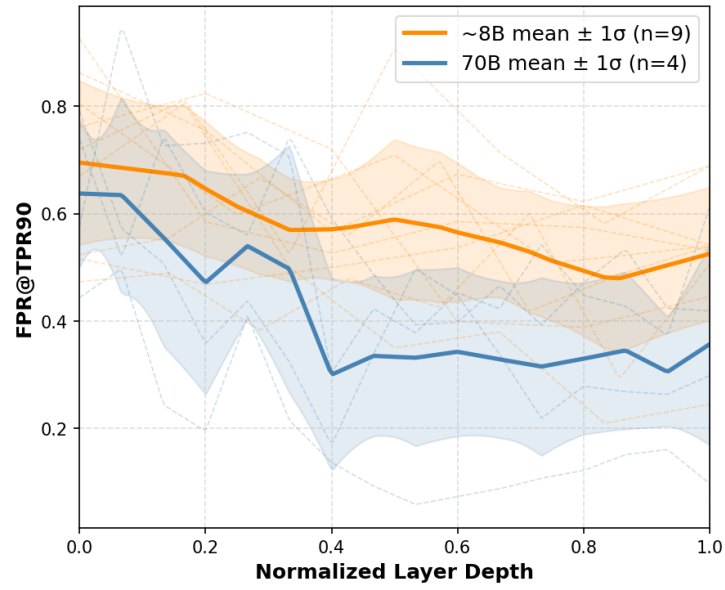}
    \caption {FPR@90\%TPR plotted against normalized probing layer depth. Lower is better. Solid line: mean. Shaded area: $\pm 1\sigma$.}\label{fig:fpr_at_tpr}
\end{figure}

We measure FPR@90\%TPR in order to characterize the operational cost of monitoring models at a set score threshold for a probe. We again see significant differences in this metric when comparing 70B models versus $\sim$8B models. For the 70B class of models, the average FPR@90\%TPR in later layers is approximately 34\% (std. ~14\%), while for the 8B class the average is between 50-60\% (std. ~10\%). In summary, these results suggest that probes operating on larger models achieve lower operational cost at a set threshold, as measured by FPR@90\%TPR.

%%%%%%%%%%%%%%%%%%%%%%%%%%%%%%%%%%%%%%%%%%%%%%%%%%%%%%%%%%%%

% \newpage
% \input{checklist.tex}

\end{document}